\documentclass{article}
\usepackage{iclr2027_conference,times}

\usepackage{amsmath,amsfonts,bm}

\def\eqref#1{equation~\ref{#1}}

\def\1{\bm{1}}

\DeclareMathAlphabet{\mathsfit}{\encodingdefault}{\sfdefault}{m}{sl}
\SetMathAlphabet{\mathsfit}{bold}{\encodingdefault}{\sfdefault}{bx}{n}

\usepackage{hyperref}
\usepackage{url}
\usepackage{graphicx}
\usepackage{float}
\usepackage{wrapfig}
\usepackage{needspace}

\usepackage{xcolor}
\usepackage{colortbl}

\newcommand{\methodname}{VR-JEPA}

\title{\methodname{}: Learning Contrastive-State Latent Guidance for Generation-based Video Reasoning}

\author{\begin{minipage}[t]{5.25in}\raggedright
{\large\bfseries
Zehua Ma\textsuperscript{1*}, Kun Xiang\textsuperscript{1*},
Yunshuang Nie\textsuperscript{3}\\
Quanlin Chen\textsuperscript{1,2}, Haoyuan Li\textsuperscript{1},
Xiuwei Chen\textsuperscript{1}\\
Jiang Ji\textsuperscript{4}, Haijun Wu\textsuperscript{4},
Zhenyu Xie\textsuperscript{5}\\
Michael Kampffmeyer\textsuperscript{6},
Hanhui Li\textsuperscript{1$\dagger$},
Xiaodan Liang\textsuperscript{1$\dagger$}\endgraf}
\vspace{0.5em}
{\normalfont\small
\textsuperscript{1}Shenzhen Campus of Sun Yat-sen University\\
\textsuperscript{2}Shenzhen Loop Area Institute\\
\textsuperscript{3}Tsinghua Shenzhen International Graduate School\\
\textsuperscript{4}Tencent\\
\textsuperscript{5}Mohamed bin Zayed University of Artificial Intelligence\\
\textsuperscript{6}UiT The Arctic University of Norway\\[0.3em]
\texttt{mazh58@mail2.sysu.edu.cn}}
\end{minipage}}

\usepackage{pifont}
\usepackage{booktabs}
\usepackage{multirow}
\usepackage{tabularx}
\usepackage{array}

\newcommand{\cmark}{\ding{51}}%
\newcommand{\xmarkg}{\textcolor{lightgray}{\ding{55}}}%

\iclrfinalcopy 
\begin{document}

\maketitle
\begingroup
\renewcommand{\thefootnote}{}
\footnotetext{\begin{tabular}[t]{@{}l@{\,}l@{}}
\textsuperscript{*} & Both authors contributed equally.\\
\textsuperscript{$\dagger$} & Corresponding author.
\end{tabular}}
\endgroup

\begin{figure}[H]
    \centering
    \includegraphics[width=\textwidth,height=0.34\textheight,keepaspectratio]{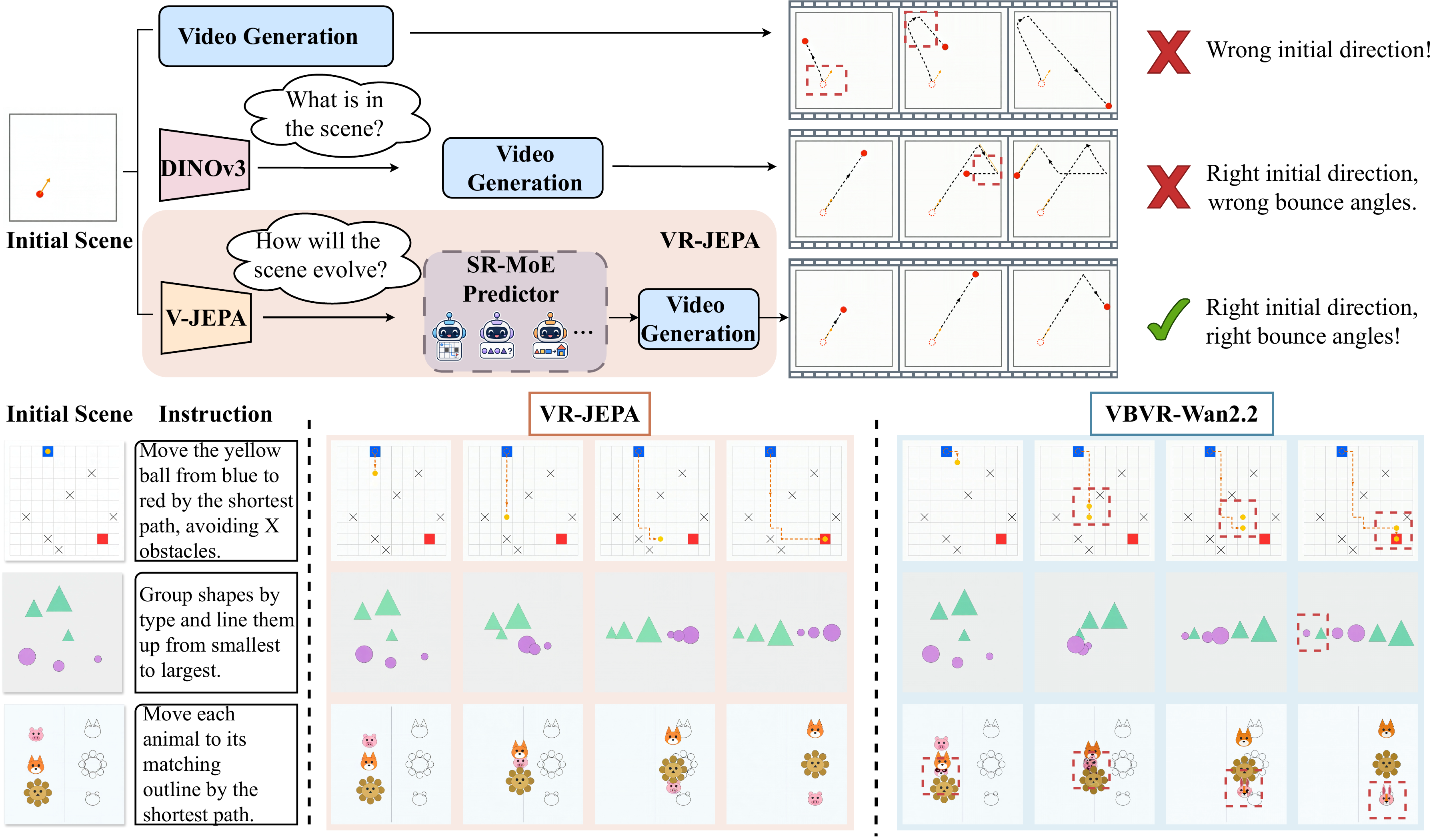}
    \caption{\textbf{Generation-based visual reasoning with \methodname{}.}  Top: Comparison of different video reasoning paradigms: (1) direct video generation; (2) video generation guided by static DINOv3 features of the initial state; and (3) VR-JEPA, which adapts V-JEPA’s spatiotemporal priors to diverse reasoning skills and guides video generation through predicted latent trajectories. Bottom: Video reasoning results of VR-JEPA and VBVR-Wan2.2, showing that VR-JEPA generates trajectories that satisfy task requirements. Red dashed boxes highlight reasoning errors.
    }
    \label{fig:teaser}
\end{figure}


\begin{abstract}

Reasoning through video generation offers a promising path toward visual intelligence by modeling latent visual states and their dynamics. However, current video generation models often lack explicit guidance on how these states should evolve, leaving generated trajectories prone to physical and structural inconsistencies that undermine reasoning reliability. While the Video Joint-Embedding Predictive Architecture (V-JEPA) provides rich spatiotemporal priors learned through latent prediction, these general priors do not naturally adapt to the logical reasoning capabilities required for complex visual tasks. To bridge this gap, we propose VR-JEPA, a framework that aligns the V-JEPA predictor with task-specific reasoning logic through localized contrastive-state learning and uses its predicted latent trajectories to guide video generation for visual reasoning. Specifically, (i) we pair successful trajectories with generated alternatives under the same input conditions and use discrepancies in their V-JEPA representations to identify informative states and tokens for localized contrastive supervision. (ii) We further equip the V-JEPA predictor with skill-specific experts trained on anchor-task data, allowing the model to adaptively specialize its shared spatiotemporal priors across diverse cognitive domains.
Together with skill-specific experts, this contrastive supervision enables VR-JEPA to predict latent trajectories that provide task-specific logical guidance for video generation. Comprehensive experiments on the large-scale VBVR-Pro-Bench dataset demonstrate that VR-JEPA achieves an $11.33\%$ relative improvement over the cutting-edge generation-based reasoning baseline, significantly mitigating physical artifacts and enhancing logical consistency.
\end{abstract}

\section{Introduction}

Anticipating how the world will change is central to planning and problem solving, motivating world models that support prediction and decision making \citep{ha2018worldmodels,micheli2023iris,du2024videolanguageplanning,xiang2025physicalai}. Video generation provides a setting for examining predicted scene evolution: it renders intermediate state transitions and final outcomes as a sequence. Diffusion models have made substantial progress in generating realistic and temporally coherent videos \citep{ho2022videodiffusion,blattmann2023videoldm}.
However, their denoising objective does not by itself ensure that generated transitions satisfy a task's requirements. A video may look convincing while moving the wrong object or depicting an incorrect object relation. This gap motivates guidance that links generated state changes to task requirements~\citep{liu2026wanr1,cheng2026vlmteachers}.

The Video Joint-Embedding Predictive Architecture (V-JEPA) offers pretrained spatiotemporal representations that can support such guidance. By predicting visual features rather than pixels, it captures scene content and motion while reducing emphasis on appearance details \citep{bardes2024vjepa}. Yet representing how a scene evolves is not the same as predicting how it should evolve to satisfy an instruction. A predictor may track an object's motion while missing the spatial relation that motion must establish. Furthermore, different tasks also require distinct reasoning skills, such as understanding physical interactions, comparing quantities, and inferring abstract rules. A predictor built on V-JEPA features must therefore focus on task-relevant transitions and adapt to diverse reasoning requirements. Its predicted trajectories should guide video generation toward sequences that satisfy these requirements.
\par

To address these challenges, we introduce \emph{\methodname{}}, a framework that predicts task-relevant trajectories in a latent space derived from V-JEPA features and uses them to guide video generation (Figure~\ref{fig:teaser}). Given an initial scene and an instruction, a learned predictor autoregressively produces a sequence of compact latent states describing the desired scene evolution. These states condition a video diffusion model, separating the prediction of task-relevant states from visual synthesis.


\par

\par
Specifically, to make latent prediction more sensitive to task-relevant changes, \methodname{} trains the predictor with localized contrastive-state learning. A solution may depend on only a few object movements or changes in inter-object relations, while many other visual details are incidental. We pair successful reference trajectories with task-matched generated candidates and use discrepancies in their V-JEPA features as a heuristic for weighting contrastive supervision at the state and token levels. This encourages predicted rollouts to follow the reference evolution rather than merely capture generic scene motion. Complementing this focused supervision, \methodname{} incorporates a \emph{skill-routed mixture of experts} (SR-MoE) to model heterogeneous state transitions. Lightweight residual experts are first trained on representative anchor tasks; a sparse, state-conditioned router then learns to combine them as the rollout evolves, using the predictor's training objectives without expert-assignment labels. 
\par

On VBVR-Pro-Bench, \methodname{} improves overall task success from 50.3\% for VBVR-Wan2.2 to 56.0\%. Component ablations support the contributions of localized contrastive-state learning and skill routed mixture of experts. Our contributions are threefold:
\begin{itemize}
    \item We introduce \methodname{}, a framework that predicts task-relevant latent trajectories from V-JEPA features and uses them to guide video generation, separating state evolution prediction from visual synthesis.

    \item We develop localized contrastive-state learning (LCL), which uses feature discrepancies between successful reference trajectories and task-matched generated candidates to focus contrastive supervision on informative states and tokens.
    
    \item We develop a skill-routed mixture of experts (SR-MoE) for latent prediction. Anchor-task-trained residual experts are combined by a sparse, state-conditioned router to adapt to heterogeneous state transitions without expert-assignment labels.
\end{itemize}


\begin{figure*}[t]
    \centering
    \includegraphics[width=\textwidth]{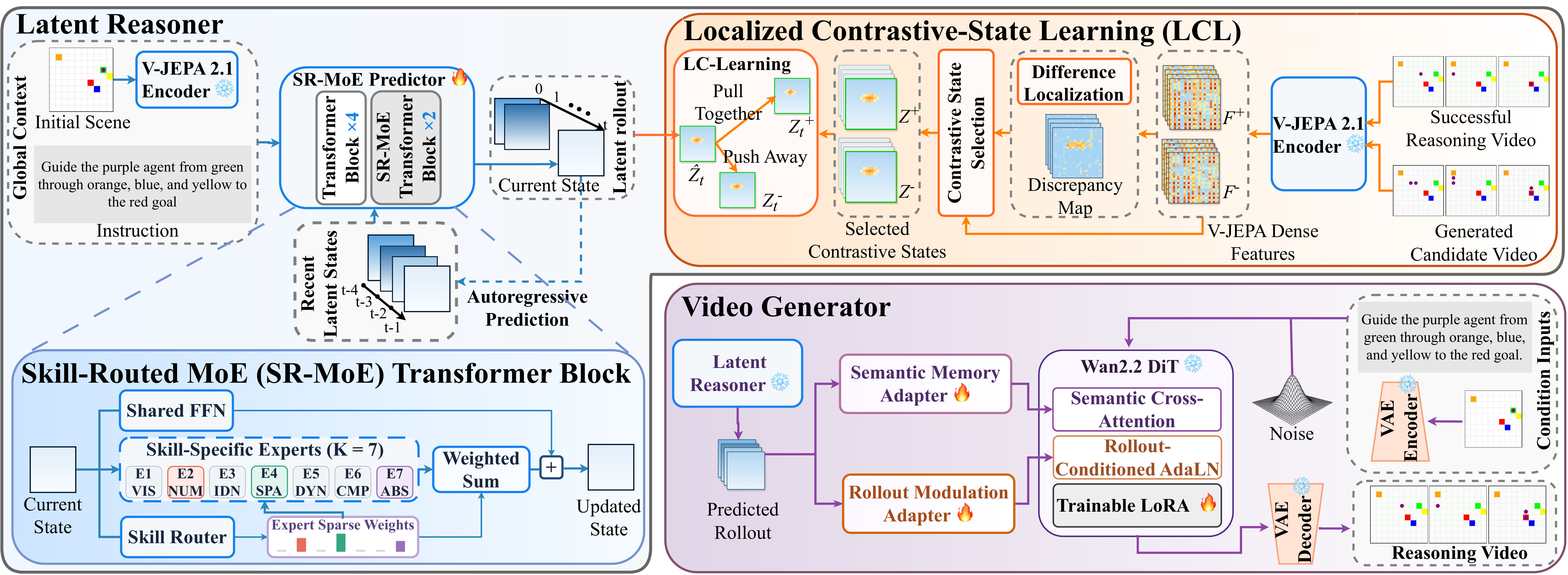}
    \vspace{-4mm}
    \caption{\textbf{Overview of \methodname{}.} \methodname{} comprises a Latent Reasoner and a Video Generator. The Latent Reasoner predicts a latent rollout from the initial scene and task instruction, with SR-MoE adaptively selecting expert combinations for different reasoning tasks. Its seven experts cover VIS (visual discrimination), NUM (numerical reasoning), IDN (object identity and persistence), SPA (spatial reasoning and planning), DYN (physical dynamics), CMP (compositional reasoning), and ABS (abstract rule inference). During predictor training, localized contrastive-state learning uses dense V-JEPA discrepancies between successful and generated videos to focus contrastive supervision on tokens corresponding to informative states and regions. The Video Generator conditions Wan2.2 DiT on the rollout to synthesize a reasoning video.}
    \label{fig:method_overview}
\end{figure*}

\section{Related Work}

\subsection{Joint-Embedding Predictive Architectures (JEPA)}


Joint-Embedding Predictive Architecture (JEPA) learns representations by predicting target embeddings from contextual observations.
I-JEPA introduces this approach for images, and V-JEPA extends it to spatiotemporal representations \citep{assran2023ijepa,bardes2024vjepa}.
V-JEPA 2 demonstrates latent prediction for video understanding and action-conditioned planning \citep{assran2025vjepa2}.
Recent work extends JEPA to reasoning and generation.
JEPA-Reasoner separates latent reasoning from text generation, while ThinkJEPA incorporates vision-language reasoning features into latent world modeling \citep{liu2025jepareasoner,zhang2026thinkjepa}.
For video generation, WMReward uses V-JEPA prediction consistency for candidate selection and sampling guidance, while Off-Manifold Refinement refines generation trajectories with prediction-energy gradients \citep{yuan2026wmreward,nguyentruong2026omr}.
PhysVideoGenerator predicts V-JEPA features from noisy diffusion latents and injects them into the generator \citep{satish2026physvideogenerator}.
These generation approaches primarily target physical plausibility.
\methodname{} focuses on visual problem solving, adapting a JEPA predictor through localized contrastive-state learning and skill specialization to produce latent trajectories that condition video generation.

\subsection{Visual Reasoning through Generation}

Visual generation provides a medium for expressing intermediate states during problem solving.
\citet{wiedemer2025videoreasoners} demonstrate emergent zero-shot perception, manipulation, and reasoning capabilities in video models.
Thinking with Video further examines generation-based reasoning through VideoThinkBench, covering both vision-centric and text-centric tasks \citep{tong2025thinkingvideo}.
Moving toward systematic training and evaluation, VBVR introduces large-scale reasoning data, verifiable scoring, and scaling studies with VBVR-Wan2.2 \citep{wang2026vbvr}.
VBVR-Pro extends this direction to native visual reasoning across generation modalities and evaluates transfer to external benchmarks \citep{xu2026vbvrpro}.
Beyond establishing these capabilities, recent approaches improve video reasoning through explicit feedback.
Wan-R1 studies verifiable reinforcement learning for spatial reasoning and planning, while \emph{VLMs are Good Teachers} uses a vision-language model to formulate rewards for test-time optimization of a video generator \citep{liu2026wanr1,cheng2026vlmteachers}.
These approaches motivate supervision that evaluates task correctness beyond visual plausibility.
Our work takes a complementary route by learning reasoning guidance within a latent predictor.
Rather than applying correctness feedback only to the generator, \methodname{} uses successful and contrastive trajectories to train intermediate state predictions, which subsequently guide the generation of visual solutions.

\section{Method}
\label{sec:method}

\subsection{Framework Overview}
\label{sec:overview}

As shown in Figure~\ref{fig:method_overview}, \methodname{} comprises a \emph{Latent Reasoner} and a \emph{Video Generator}. A frozen V-JEPA encoder and a learnable pooler encode the initial scene \(x_0\) as a latent state \(Z_0\). Given \(Z_0\), the instruction \(c\), and up to \(H\) recent states, the predictor \(P_\theta\) produces a latent rollout that conditions the video generator \(G_\phi\):
\begin{equation}
\widehat Z_{t+1}
=
P_\theta\!\left(Z_0,\widehat Z_{\max(0,t-H+1):t},c\right),
\qquad
\widehat{\mathcal Z}=(Z_0,\widehat Z_1,\ldots,\widehat Z_T),
\qquad
\widehat V=G_\phi(x_0,c,\widehat{\mathcal Z}).
\label{eq:framework}
\end{equation}
Here, \(\widehat Z_0=Z_0\), \(T\) is the prediction horizon, \(t=0,\ldots,T-1\), and \(H\) is the maximum number of recent predicted states supplied to the predictor. 

The rollout conditions the generator through two pathways. A \emph{semantic memory adapter} applies temporal self-attention to rollout states and projects them into keys and values, which the DiT video tokens query through gated cross-attention. A \emph{rollout modulation adapter} adds rollout-dependent residual offsets to the shift, scale, and residual-gate parameters of the DiT self-attention and MLP branches, following the adaptive modulation used in DiT~\citep{peebles2023dit}.

\subsection{Localized Contrastive-State Learning}
\label{sec:counterfactual}

V-JEPA's general spatiotemporal priors may not emphasize the state transitions and regions critical to visual reasoning.
To address this limitation, \methodname{} introduces localized contrastive-state learning (LCL), which compares successful videos with task-matched generated candidates in dense V-JEPA feature space, identifies informative states and regions, and strengthens contrastive supervision on the corresponding latent tokens.

\paragraph{V-JEPA latent rollout prediction.}
For each successful training video, frozen V-JEPA 2.1 encoder \(E\)~\citep{murlabadia2026vjepa21} extracts features \(F_t^+\) from local clip \(v_t^+\). The learnable pooler \(Q_\psi\) maps them to \(M\) tokens of dimension \(d\):
\begin{equation}
F_t^+=E(v_t^+), \qquad
Z_t^+=Q_\psi(F_t^+)\in\mathbb{R}^{M\times d}.
\label{eq:state_representation}
\end{equation}
We train the pooler and predictor with one- and two-step prediction (feeding back the first prediction) and VISReg regularization~\citep{assran2025vjepa2,wu2026visreg}:
\begin{equation}
\begin{gathered}
\mathcal L_j=
\frac{1}{|\mathcal T_j|Md}
\sum_{t\in\mathcal T_j}
\left\|\nu(\widehat Z_{t+j\mid t})
-\operatorname{sg}\!\left[\nu(Z^+_{t+j})\right]\right\|_F^2,
\quad j\in\{1,2\},\\
\mathcal L_{\mathrm{base}}
=\mathcal L_1+\lambda_2\mathcal L_2
+\lambda_{\mathrm{reg}}\mathcal R_{\mathrm{VISReg}}.
\end{gathered}
\label{eq:prediction_loss}
\end{equation}
Here, \(\widehat Z_{t+j\mid t}\) is the \(j\)-step prediction from ground-truth history at \(t\); \(\mathcal T_j\) contains valid starting times. The operator \(\nu\) applies token-wise \(\ell_2\) normalization, and \(\operatorname{sg}\) stops gradients through the target. The coefficients \(\lambda_2\) and \(\lambda_{\mathrm{reg}}\) weight the two-step and regularization losses.

\par\Needspace{0.65\textheight}
\begin{wrapfigure}{R}{0.6\textwidth}
    \centering
    \includegraphics[width=\linewidth]{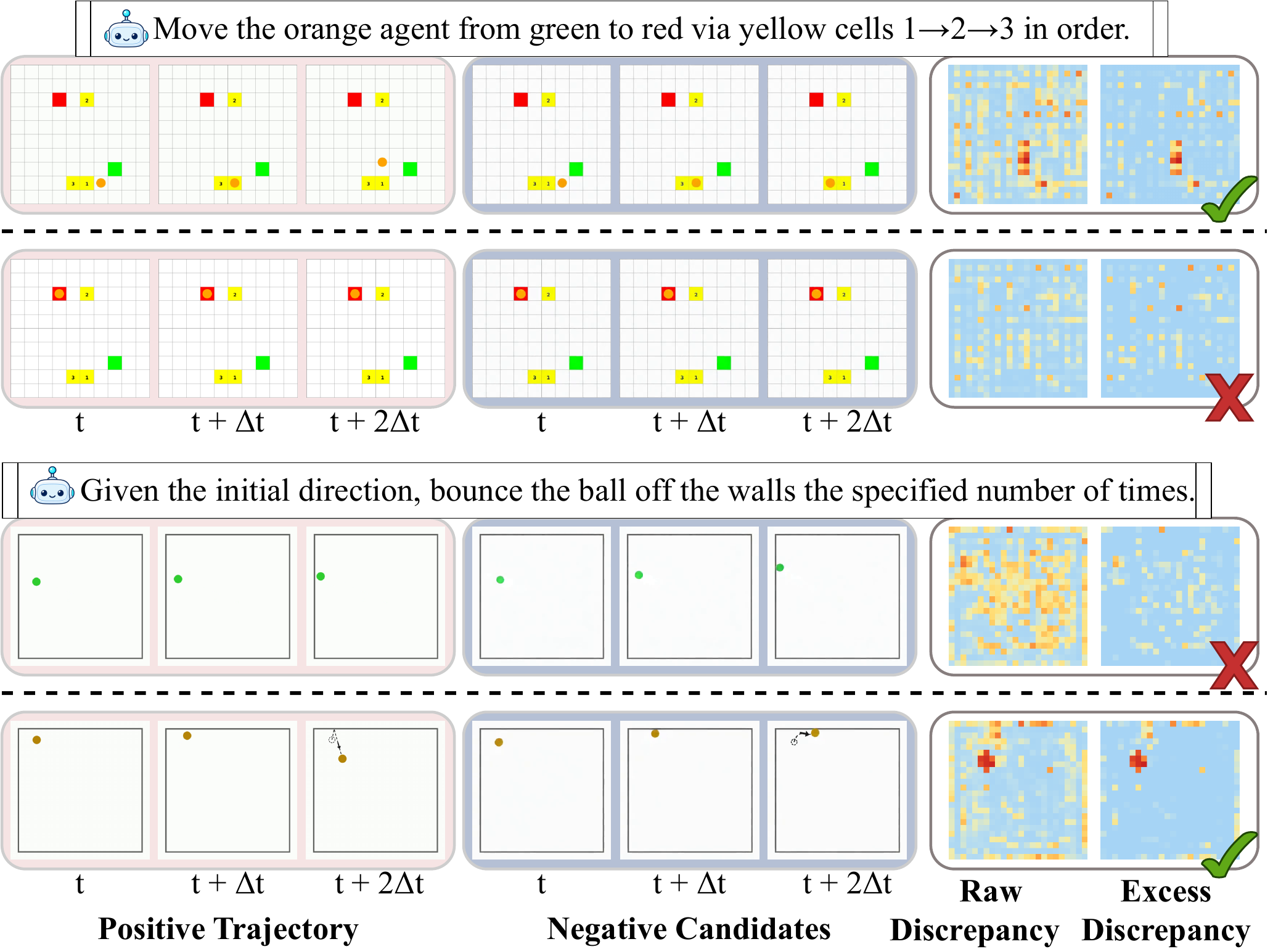}
    \vspace{-2mm}
    \caption{
    \textbf{State selection for localized contrastive-state learning.} Successful reference trajectories and task-matched generated candidates are compared in dense V-JEPA feature space. After state-specific quantile subtraction, large residual discrepancies select states for contrastive supervision (green checks), while small discrepancies exclude them (red crosses).
    }
    \label{fig:localized_counterfactual}
\end{wrapfigure}

\paragraph{Trajectory differences localization.}

\methodname{} treats each training video $V^+$ as a successful reasoning sample and generates a candidate video $V^-$ using VBVR-Wan2.2~\citep{wang2026vbvr} under the same input conditions.
Since the generated candidate may satisfy the reasoning requirements, it may not serve as a valid negative sample.
\methodname{} uses discrepancies between their dense V-JEPA features as a heuristic to select pairs and locate states and regions that merit stronger supervision.
Figure~\ref{fig:localized_counterfactual} shows the selection and~\mbox{localization}.

To measure differences between each training video and its candidate, the frozen V-JEPA encoder processes clips at the same temporal index $t$, yielding $F_t^\pm=E(v_t^\pm)$.
\methodname{} computes cosine discrepancies at corresponding feature-grid positions $n$.
To reduce the influence of scene-wide appearance differences, it subtracts the $q$-quantile of discrepancies within each state and retains the positive residuals:
\begin{equation}
D_{t,n} = 1-\frac{\langle F^+_{t,n},F^-_{t,n}\rangle}{\|F^+_{t,n}\|_2\|F^-_{t,n}\|_2},
\qquad
S_{t,n} = \left[D_{t,n}-\operatorname{Quantile}_{q}(D_{t,:})\right]_+.
\label{eq:localized_discrepancy}
\end{equation}

Here, $q$ is the quantile-level hyperparameter, and $[x]_+=\max(x,0)$.
Averaging the residuals over the $N$ dense positions yields the state score $s_t=N^{-1}\sum_n S_{t,n}$.
\methodname{} maps this score to a state-level weight $a_t=\operatorname{clip}((s_t-\tau_{\min})/(\tau_{\mathrm{full}}-\tau_{\min}),0,1)$, which scales the state's contribution to the contrastive-state loss.
The thresholds $\tau_{\min}$ and $\tau_{\mathrm{full}}$ specify the scores at which this weight is zero and one, respectively.

For each state with $a_t>0$, \methodname{} uses the pooler's attention maps to transfer localized discrepancies to latent tokens.
Let $A^+_{t,h,m,n}$ denote attention from token $m$ to dense position $n$ in head $h$ for the successful video.
The normalized discrepancy map and token weights are computed as
\begin{equation}
\widetilde S_{t,n}=\frac{S_{t,n}}{\sum_{n'}S_{t,n'}+\epsilon},
\qquad
b_{t,m}=\frac{1}{N_h}\sum_h
\left[\sum_n A^+_{t,h,m,n}\widetilde S_{t,n}-\frac{1}{N}\right]_+,
\label{eq:token_localization}
\end{equation}

where $N_h$ is the number of attention heads, and $\epsilon>0$ ensures numerical stability.
The term $1/N$ provides a uniform-attention baseline over the $N$ dense positions.
A token receives positive weight only if its attention to the localized discrepancies exceeds this baseline in at least one head; states with no positively weighted tokens are excluded.

\paragraph{Contrastive-state supervision.}

Using the state and token weights, \methodname{} encourages each one-step prediction to be closer to the successful target than to its paired generated candidate.
Let $\Omega$ denote the selected state--token positions with $a_t>0$ and $b_{t,m}>0$.
The weighted margin loss and overall training objective are
\begin{equation}
\begin{gathered}
\mathcal L_{\mathrm{LCL}}
=
\frac{1}{|\Omega|}
\sum_{(t,m)\in\Omega}a_t b_{t,m}
\left[\mu+d(\widehat Z_{t,m},Z^+_{t,m})-d(\widehat Z_{t,m},Z^-_{t,m})\right]_+,\\
\mathcal L_{\mathrm{reason}}
=
\mathcal L_{\mathrm{base}}+\lambda_{\mathrm{LCL}}\mathcal L_{\mathrm{LCL}}.
\end{gathered}
\label{eq:counterfactual_loss}
\end{equation}

Here, $\widehat Z_{t,m}$ denotes token $m$ of the one-step prediction $\widehat Z_{t\mid t-1}$, while $Z^+_{t,m}$ and $Z^-_{t,m}$ are the corresponding pooled tokens from the successful and candidate videos.
The function $d$ is cosine distance, $\mu$ is the margin, and $\lambda_{\mathrm{LCL}}$ controls the contribution of contrastive-state supervision.
Gradients are stopped through the weights and both target representations, and $\mathcal L_{\mathrm{LCL}}=0$ when $\Omega$ is empty.

\subsection{Skill-Routed Mixture of Experts}
\label{sec:experts}

A compact V-JEPA predictor may lack the capacity to model the diverse reasoning patterns required across tasks. To address this limitation, \methodname{} equips the predictor with a lightweight skill-routed mixture of experts, which adaptively selects different expert combinations for different tasks.


\paragraph{Learning skill-specific experts.}

\methodname{} first trains experts on representative anchor tasks to encourage specialization and provide a skill-oriented basis for subsequent composition.
Starting from the trained dense predictor, SR-MoE adds lightweight bottleneck adapters to the final two blocks of the six-layer predictor while retaining the shared feed-forward networks:
\begin{equation}
Y^{(\ell)}_{t,m}
=
F^{(\ell)}_{\mathrm{shared}}\!\left(H^{(\ell)}_{t,m}\right)
+
\alpha\sum_{k=1}^{K}\pi^{(\ell)}_{t,k}A^{(\ell)}_k\!\left(H^{(\ell)}_{t,m}\right).
\label{eq:expert_prediction}
\end{equation}

Here, $H^{(\ell)}_{t,m}$ is the input token, $Y^{(\ell)}_{t,m}$ is its expert-augmented output, $F^{(\ell)}_{\mathrm{shared}}$ is the shared feed-forward network, and $A_k^{(\ell)}$ is the $k$-th expert adapter.
The expert residual coefficient is fixed to $\alpha=1$, so the weighted expert output is added directly to the shared output without an additional learned or scheduled scale; $\pi^{(\ell)}_{t,k}$ is the routing weight.
The model uses $K=7$ skill-specific experts. Please refer to Figure~\ref{fig:method_overview} for their abbreviations and corresponding skills.

For each anchor task, \methodname{} directly activates the expert associated with its skill and updates only that expert's adapters, leaving the shared predictor fixed.
This gives each expert an initial skill-specific capability before the router learns to combine them across tasks.

\paragraph{Learning adaptive routing.}

Different categories of reasoning tasks require different skills or combinations of skills.
\methodname{} therefore learns a router to select and combine the pretrained experts according to each task's reasoning requirements.
The router conditions on the current state, the preceding state, their difference, and the task context:

\begin{equation}
r_t^{(\ell)}
=
\left[\bar H_t^{(\ell)};\bar H_{t-1}^{(\ell)};\bar H_t^{(\ell)}-\bar H_{t-1}^{(\ell)};g^{(\ell)}\right],
\qquad
\pi_t^{(\ell)}
=
\operatorname{Entmax}_{1.5}\!\left(f_{\mathrm{route}}^{(\ell)}(r_t^{(\ell)})\right).
\label{eq:expert_routing}
\end{equation}

Here, $\bar H_t^{(\ell)}$ averages the hidden tokens within a state, $g^{(\ell)}$ summarizes the conditioning context, and $f_{\mathrm{route}}^{(\ell)}$ maps the concatenated features $r_t^{(\ell)}$ to expert scores.
$\operatorname{Entmax}_{1.5}$~\citep{peters2019sparse} converts these scores into normalized sparse weights, enabling the router to select and combine experts for different reasoning tasks.

The router is trained through the prediction and contrastive-state objectives, without expert-assignment annotations for the full training set or additional load-balancing losses.
Training first optimizes the router with fixed experts, then jointly adapts the router and experts, and finally updates selected shared components.
This progression learns how to combine the initialized experts before refining their capabilities and coordination beyond the anchor tasks.
The resulting adaptive composition supports latent rollout prediction for tasks requiring multiple reasoning skills. 

\section{Experiments}



\subsection{Experimental Setup}

\paragraph{Benchmark and Evaluation.}
We evaluate \methodname{} on VBVR-Pro-Bench~\citep{xu2026vbvrpro}, which comprises 100 visual reasoning tasks spanning abstraction, knowledge, perception, spatial reasoning, and transformation. We evaluate five instances per task, totaling 500 instances. Given an initial image and an instruction, models generate a video depicting the solution. The benchmark includes 50 in-domain (ID) tasks from task families represented in its training set and 50 held-out out-of-domain (OOD) tasks. Task-specific, rule-based evaluators assess generated outputs against the required outcomes and constraints. We report overall, ID/OOD, and category-level scores; absolute gains are expressed in percentage points.

The publicly released baseline outputs for VBVR-Pro-Bench differ in frame count and spatial resolution. We rerun VBVR-Wan2.2 on VBVR-Pro-Bench. Before evaluation, we temporally sample each generated video to match the frame count of its corresponding ground-truth video and resize the frames to $512\times512$ pixels. We apply this preprocessing to outputs from all compared methods and recompute their scores using the same evaluation pipeline, ensuring a consistent comparison.

\paragraph{Baselines.}
We compare \methodname{} with six open-source video models:
Wan2.2-I2V-A14B and Wan2.2-TI2V-5B~\citep{wan2025wan22},
Wan2.1-I2V-14B-720P~\citep{wan2025},
LTX-2.3-I2AV~\citep{hacohen2026ltx2,lightricks2026ltx23},
CogVideoX1.5-5B-I2V~\citep{yang2024cogvideox,cogvideox15},
and HunyuanVideo-I2V~\citep{kong2024hunyuanvideo,tencent2025hunyuanvideoi2v}.
We also include three proprietary video models:
Seedance 2.0~\citep{teamseedance2026},
Kling VIDEO 3.0~\citep{kling2026guide}, 
and Veo 3.1~\citep{gallegos2025veo31}.
Among video reasoning models, our primary baseline is VBVR-Wan2.2~\citep{wang2026vbvr}, which adapts Wan2.2-I2V-A14B to the VBVR dataset without modifying the model architecture.

\paragraph{Implementation Details.}
\methodname{} is trained in two stages. First, the Latent Reasoner is trained with the V-JEPA 2.1 encoder~\citep{murlabadia2026vjepa21} frozen. The learnable pooler and predictor are optimized using the prediction and localized contrastive-state objectives described in Section~\ref{sec:counterfactual}, with expert training and routing detailed in Section~\ref{sec:experts}. Second, the Latent Reasoner is frozen, and the Wan2.2-14B generator, initialized from VBVR-Wan2.2, is adapted by optimizing only the LoRA parameters~\citep{hu2022lora} and rollout-conditioning modules with a flow-matching loss~\citep{lipman2023flowmatching}. Additional implementation details are provided in Appendix~\ref{app:implementation}.

\begin{table*}[t]
    \centering
    \caption{
        \textbf{Main results on VBVR-Pro-Bench~\citep{xu2026vbvrpro}.}
        We report overall, in-domain (ID), and out-of-domain (OOD) scores, followed by results for abstraction (Abst.), knowledge (Know.), perception (Perc.), spatial reasoning (Spat.), and transformation (Trans.).
        Avg. denotes the mean of task scores weighted by the number of samples per task.
        All scores are reported as percentages. Higher is better.
        Bold and underlined values indicate the best and second-best results.
    }
    \label{tab:main_results}

    \fontsize{7.5pt}{9pt}\selectfont
    \setlength{\tabcolsep}{1.5pt}
    \renewcommand{\arraystretch}{1.15}

    \begin{tabular*}{\textwidth}{
        @{\extracolsep{\fill}}
        >{\raggedright\arraybackslash}p{0.22\textwidth}|
        c|
        c|*{5}{c}|
        c|*{5}{c}
        @{}
    }
        \toprule
        &
        & \multicolumn{6}{c|}{\textbf{In-Domain (ID) (\%)}}
        & \multicolumn{6}{c}{\textbf{Out-of-Domain (OOD) (\%)}} \\
        \cmidrule(lr){3-3}
        \cmidrule(lr){4-8}
        \cmidrule(lr){9-9}
        \cmidrule(lr){10-14}

        \textbf{Models} & \textbf{Overall (\%)} & \textbf{Avg.}
          & \textbf{Abst.}
          & \textbf{Know.}
          & \textbf{Perc.}
          & \textbf{Spat.}
          & \textbf{Trans.}
          & \textbf{Avg.}
          & \textbf{Abst.}
          & \textbf{Know.}
          & \textbf{Perc.}
          & \textbf{Spat.}
          & \textbf{Trans.} \\
        \midrule

        \rowcolor[RGB]{243,248,252}
        \multicolumn{14}{l}{\textbf{Open-source Video Models}} \\
        Wan2.2-I2V-A14B & 18.2 & 15.7 & 12.3 & 15.2 & 15.0 & 17.0 & 23.3 & 20.7 & 29.2 & 17.9 & 13.8 & 19.5 & 31.8 \\
        LTX-2.3-I2AV & 11.2 & 10.6 & 9.3 & 12.3 & 9.7 & 13.6 & 5.1 & 11.9 & 21.7 & 19.8 & 7.0 & 9.1 & 5.9 \\
        Wan2.1-I2V-14B-720P & 10.0 & 10.5 & 8.7 & 14.6 & 12.4 & 9.9 & 4.0 & 9.5 & 13.9 & 8.7 & 7.6 & 12.3 & 5.2 \\
        Wan2.2-TI2V-5B & 9.4 & 6.6 & 5.0 & 7.4 & 7.7 & 9.2 & 2.5 & 12.2 & 22.2 & 9.2 & 9.1 & 6.3 & 11.6 \\
        CogVideoX1.5-5B-I2V & 8.6 & 9.7 & 9.2 & 14.2 & 7.8 & 7.9 & 5.5 & 7.5 & 16.5 & 5.6 & 5.0 & 3.9 & 2.8 \\
        HunyuanVideo-I2V & 5.6 & 5.5 & 4.0 & 7.2 & 1.0 & 9.1 & 3.8 & 5.7 & 11.6 & 1.5 & 2.7 & 7.4 & 6.4 \\
        \midrule

        \rowcolor[RGB]{243,248,252}
        \multicolumn{14}{l}{\textbf{Proprietary Video Models}} \\
        Seedance 2.0 & 48.5 & 43.9 & \underline{41.9} & 40.6 & 47.5 & 49.7 & 41.8 & \textbf{53.0} & \textbf{43.7} & \underline{67.9} & \textbf{52.4} & 53.3 & 61.3 \\
        Kling VIDEO 3.0 & 38.7 & 36.1 & 28.7 & 34.6 & 48.2 & 40.3 & 34.6 & 41.3 & 34.6 & \textbf{75.5} & \underline{40.1} & 24.1 & 47.7 \\
        Veo 3.1 & 14.9 & 16.3 & 15.1 & 20.5 & 19.6 & 13.6 & 9.7 & 13.4 & 21.2 & 14.8 & 10.1 & 12.9 & 9.1 \\
        \midrule

        \rowcolor[RGB]{243,248,252}
        \multicolumn{14}{l}{\textbf{Video Reasoning Models}} \\
        VBVR-Wan2.2 & \underline{50.3} & \underline{55.4} & 33.8 & \underline{54.8} & \underline{50.9} & \underline{65.9} & \underline{91.2} & 45.1 & 35.8 & 42.7 & 34.0 & \textbf{72.2} & \textbf{77.7} \\
        \textbf{VR-JEPA} & \textbf{56.0} & \textbf{65.9} & \textbf{56.3} & \textbf{63.8} & \textbf{61.8} & \textbf{69.1} & \textbf{91.4} & \underline{46.1} & \underline{41.5} & 42.1 & 35.2 & \underline{65.2} & \underline{77.5} \\

        \bottomrule
    \end{tabular*}
\end{table*}

\subsection{Main Results}

\paragraph{Quantitative Comparisons.}
Table~\ref{tab:main_results} reports quantitative comparisons on VBVR-Pro-Bench. \methodname{} achieves the highest overall score of 56.0\% among the evaluated models, outperforming VBVR-Wan2.2 by 5.7 percentage points and Seedance 2.0, the highest-scoring proprietary model overall, by 7.5 points. Compared with VBVR-Wan2.2, it improves the ID average from 55.4\% to 65.9\% and the OOD average from 45.1\% to 46.1\%. In particular, ID abstraction and perception improve by 22.5 and 10.9 points, reaching 56.3\% and 61.8\%, respectively. ID knowledge and spatial reasoning also improve by 9.0 and 3.2 points, while transformation performance remains nearly unchanged at 91.4\%. These gains over VBVR-Wan2.2 support the effectiveness of the proposed latent reasoning framework, which combines supervision on informative states and regions with adaptive skill-expert composition. Section~\ref{sec:ablations} examines the contributions of localized contrastive-state learning and SR-MoE individually and jointly.

\begin{figure}[t]
    \centering
    \includegraphics[width=\linewidth]{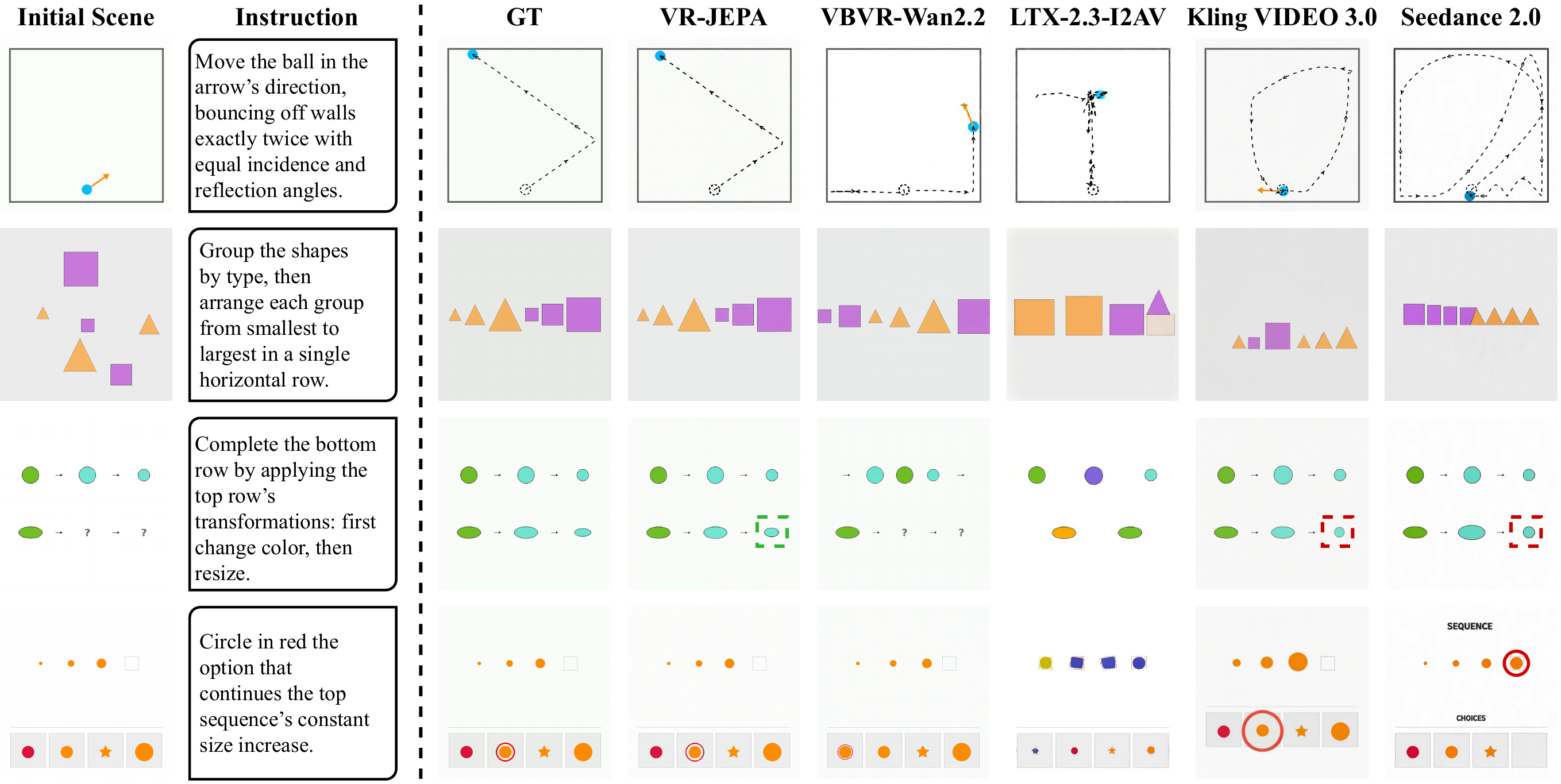}
    \vspace{-4mm}
    \caption{\textbf{Qualitative comparison on VBVR-Pro-Bench.} Given the initial state and instruction, we compare the reasoning results of \methodname{} with those of a video reasoning model (VBVR-Wan2.2), an open-source video model (LTX-2.3-I2AV), and two proprietary video models (Kling VIDEO 3.0 and Seedance 2.0). Green and red dashed boxes highlight correct results and reasoning errors, respectively.
    }
    \label{fig:qualitative_comparison}
\end{figure}


\begin{figure}[t]
    \centering
    \includegraphics[width=\linewidth]{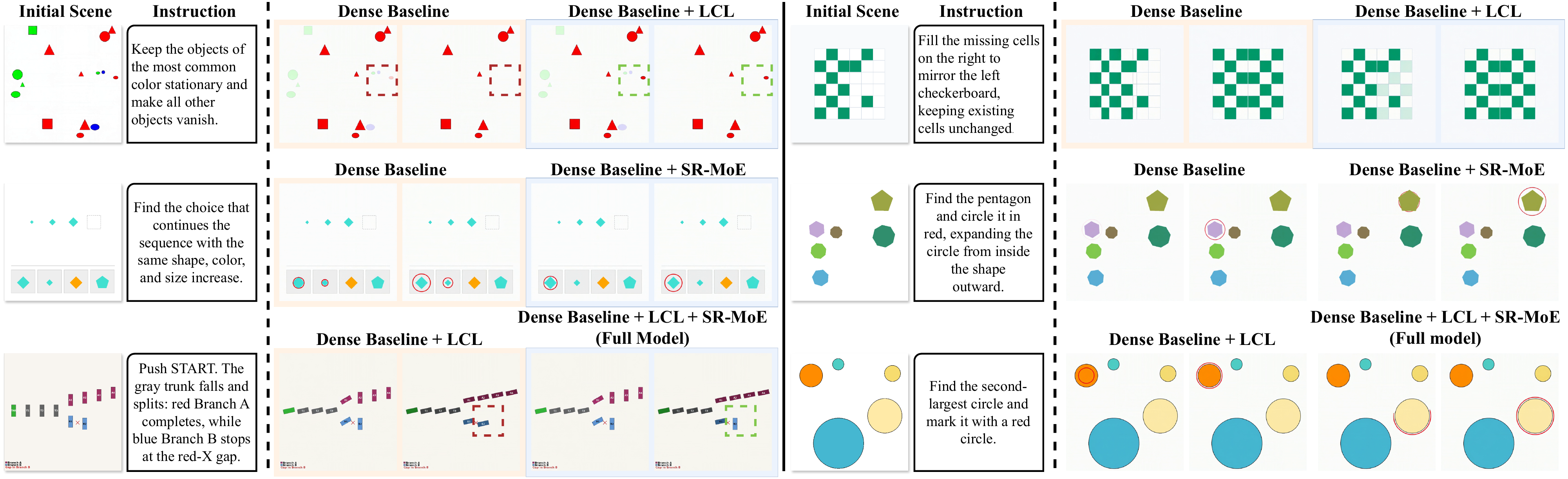}
    \vspace{-4mm}
    \caption{\textbf{Qualitative ablation of localized contrastive-state learning (LCL) and skill-routed mixture of experts (SR-MoE).} Top and middle: adding LCL and SR-MoE to the dense baseline, respectively. Bottom: adding SR-MoE to the LCL-equipped predictor to obtain the full model. Green and red dashed boxes highlight correct results and reasoning errors, respectively.}
    \vspace{-2mm}
    \label{fig:component_ablation}
\end{figure}

\paragraph{Qualitative Comparisons.}
Figure~\ref{fig:qualitative_comparison} compares \methodname{} with VBVR-Wan2.2, LTX-2.3-I2AV, Kling VIDEO 3.0, and Seedance 2.0 on four visual reasoning examples. In the ball-bouncing task, \methodname{} produces a trajectory closely matching the ground truth, whereas VBVR-Wan2.2 follows an incorrect initial direction and the other models produce curved or irregular paths. In the shape-sorting task, \methodname{} preserves the objects and groups them by type in ascending size order. By contrast, VBVR-Wan2.2 interleaves the two shape groups, while LTX-2.3-I2AV alters object shapes and colors. In the transformation task, \methodname{} correctly applies the demonstrated color change followed by size reduction to the bottom-row ellipses. VBVR-Wan2.2 leaves the answer positions unfilled, while Kling VIDEO 3.0 and Seedance 2.0 change the final ellipse into a circle. In the sequence-completion task, \methodname{} circles the same option as the ground truth, whereas VBVR-Wan2.2 circles the red distractor and Seedance 2.0 modifies the sequence instead of marking an answer option. Together, these examples highlight \methodname{}'s advantage over the compared models in following task rules while preserving relevant object attributes, yielding reasoning outputs more consistent with the intended solutions.

\subsection{Ablation Studies}
\label{sec:ablations}

\begin{table*}[t]
\begingroup
    \centering
    \caption{\textbf{Ablation of localized contrastive-state learning (LCL) and skill-routed mixture of experts (SR-MoE).} Scores are reported as percentages; higher is better.}
    \label{tab:component_ablation}
    \scriptsize
    \setlength{\tabcolsep}{2pt}
    \renewcommand{\arraystretch}{1.08}
    \resizebox{\textwidth}{!}{%
    \begin{tabular}{lcc c *{6}{c} *{6}{c}}
        \toprule
        \multirow{2}{*}{\textbf{Variant}}
        & \multicolumn{2}{c}{\textbf{Components}}
        & \multirow{2}{*}{\textbf{Overall (\%)}}
        & \multicolumn{6}{c}{\textbf{In-Domain (ID) (\%)}}
        & \multicolumn{6}{c}{\textbf{Out-of-Domain (OOD) (\%)}} \\
        \cmidrule(lr){2-3}
        \cmidrule(lr){5-10}
        \cmidrule(lr){11-16}
        & \textbf{LCL} & \textbf{SR-MoE} & &
        \textbf{Avg.} & \textbf{Abst.} & \textbf{Know.} & \textbf{Perc.} & \textbf{Spat.} & \textbf{Trans.}
        & \textbf{Avg.} & \textbf{Abst.} & \textbf{Know.} & \textbf{Perc.} & \textbf{Spat.} & \textbf{Trans.} \\
        \midrule
        Dense baseline & \xmarkg & \xmarkg & 53.9 & 63.4 & 53.8 & 58.9 & 58.9 & 68.8 & 91.4 & 44.4 & 43.3 & \textbf{44.5} & 31.9 & 59.3 & 75.3 \\
        + LCL & \cmark & \xmarkg & 55.1 & 64.9 & 52.6 & 61.2 & 61.3 & \textbf{71.6} & 93.0 & 45.3 & \textbf{44.1} & 38.5 & 33.1 & \textbf{65.4} & 75.8 \\
        + SR-MoE & \xmarkg & \cmark & 55.5 & 65.2 & 54.0 & 61.4 & 60.7 & 69.8 & \textbf{95.6} & 45.9 & 42.8 & 42.3 & 34.3 & 64.2 & 77.3 \\
        Full model & \cmark & \cmark & \textbf{56.0} & \textbf{65.9} & \textbf{56.3} & \textbf{63.8} & \textbf{61.8} & 69.1 & 91.4 & \textbf{46.1} & 41.5 & 42.1 & \textbf{35.2} & 65.2 & \textbf{77.5} \\
        \bottomrule
    \end{tabular}%
    }
  \vspace{-4mm}
\par\endgroup

\vspace{\floatsep}

\begingroup
    \centering
    \small
    \setlength{\tabcolsep}{5pt}
    \renewcommand{\arraystretch}{1.05}
    \caption{\textbf{Ablation of visual representations.} Scores are reported as percentages; higher is better.}
    \label{tab:representation_ablation}
    \resizebox{0.9\columnwidth}{!}{%
        \begin{tabular}{llcc}
            \toprule
            \textbf{Visual representation} & \textbf{Representation form}
            & \textbf{ID (\%)} $\uparrow$ & \textbf{OOD (\%)} $\uparrow$ \\
            \midrule
            DINOv3 & Initial-state features & 62.8 & 45.3 \\
            DINOv3 & Predicted latent rollout & 65.3 & 43.8 \\
            V-JEPA 2.1 (\methodname{}) & Predicted latent rollout
                & \textbf{65.9} & \textbf{46.1} \\
            \bottomrule
        \end{tabular}%
    }
    \vspace{-2mm}

\par\endgroup
\end{table*}

\paragraph{Effectiveness of LCL and SR-MoE.}
The dense baseline guides Wan2.2 video generation with latent rollouts predicted by a trained V-JEPA predictor. The three variants add localized contrastive-state learning (LCL), SR-MoE, or both to this baseline. Table~\ref{tab:component_ablation} shows that LCL and SR-MoE individually raise the overall score from 53.9\% to 55.1\% and 55.5\%, respectively. Combining both achieves the best overall, ID, and OOD averages among the ablated variants: 56.0\%, 65.9\%, and 46.1\%. Figure~\ref{fig:component_ablation} illustrates these benefits: LCL corrects the mirrored checkerboard completion, while SR-MoE selects the instructed pentagon instead of a distractor. Combining both also enables the model to identify the second-largest circle, which the LCL-only variant fails to select. Together, these results support the complementary benefits of supervision focused on informative states and regions and skill-specific expert routing in improving \methodname{}'s video reasoning.

\paragraph{Visual Representation Space.}
Table~\ref{tab:representation_ablation} compares DINOv3~\citep{simeoni2025dinov3} and V-JEPA 2.1~\citep{murlabadia2026vjepa21} as representation spaces for latent reasoning under the same framework, with initial-state DINOv3 features included as a reference. The selected DINOv3 and V-JEPA 2.1 encoders have comparable parameter counts. Using V-JEPA 2.1 for latent rollout prediction achieves video reasoning scores of 65.9\% on ID tasks and 46.1\% on OOD tasks, outperforming the DINOv3 rollout variant by 0.6 and 2.3 percentage points, respectively. These results favor V-JEPA 2.1 for video reasoning, consistent with its ability to capture spatiotemporal dynamics beyond the static visual semantics primarily encoded by DINOv3. Corresponding qualitative comparisons are provided in Figure~\ref{fig:representation_ablation} in Appendix~\ref{app:representation_ablation}.

\section{Conclusion}
We presented \methodname{}, a framework that predicts latent rollouts in V-JEPA representation space to guide video generation for visual reasoning. \methodname{} uses localized contrastive-state learning to focus contrastive supervision on informative states and regions identified from successful reference videos and task-matched generated candidates. It further equips the predictor with a skill-routed mixture of experts that adaptively combines skill-specific experts for different reasoning tasks. On VBVR-Pro-Bench, \methodname{} improves the overall score from 50.3\% to 56.0\% over VBVR-Wan2.2, an absolute gain of 5.7 percentage points. Component ablations support the complementary contributions of localized supervision and skill-expert composition. Together, these results highlight the potential of reasoning-guided latent prediction for reasoning through video generation.

\newpage
\clearpage

\section{AI Use Statement}

AI tools were used to assist with literature searches. We reviewed and selected the relevant papers and independently organized and wrote the related work section. We take full responsibility for the content and references of this paper.


\bibliography{references}
\bibliographystyle{iclr2027_conference}

\appendix
\newpage


        

\clearpage
\section*{Appendix Contents}

\providecommand{\appmissing}[1]{%
    \par\noindent
    {\color{red}\textbf{[AUTHOR TO COMPLETE:} #1\textbf{]}}
    \par
}

\begin{itemize}
    \item \hyperref[app:implementation]{\textbf{\ref*{app:implementation}\quad Implementation Details}}
    \begin{itemize}
        \item \hyperref[app:training_data]{\ref*{app:training_data}\quad Training Data}
        \item \hyperref[app:training_procedure]{\ref*{app:training_procedure}\quad Architecture and Training Procedure}
        \item \hyperref[app:configuration]{\ref*{app:configuration}\quad Eight-GPU Reproduction Configuration}
        \item \hyperref[app:parameter_overhead]{\ref*{app:parameter_overhead}\quad Parameter Overhead}
    \end{itemize}

    \item \hyperref[app:ablations]{\textbf{\ref*{app:ablations}\quad Additional Ablation Studies}}
    \begin{itemize}
        \item \hyperref[app:training_strategies]{\ref*{app:training_strategies}\quad Training Strategies}
        \item \hyperref[app:routing]{\ref*{app:routing}\quad Expert Routing Strategies}
        \item \hyperref[app:rollout_interventions]{\ref*{app:rollout_interventions}\quad Interventions on the Predicted Rollout}
    \end{itemize}

    \item \hyperref[app:qualitative]{\textbf{\ref*{app:qualitative}\quad Further Visualization}}
    \begin{itemize}
        \item \hyperref[app:qualitative_open_source]{\ref*{app:qualitative_open_source}\quad Comparisons with Open-source Models}
        \item \hyperref[app:qualitative_closed_source]{\ref*{app:qualitative_closed_source}\quad Comparisons with Proprietary Video Models}
        \item \hyperref[app:representation_ablation]{\ref*{app:representation_ablation}\quad Visual Representation Space Ablation}
        \item \hyperref[app:rollout_intervention]{\ref*{app:rollout_intervention}\quad Comparison of latent trajectory guidance}
        \item \hyperref[app:routing_visualization]{\ref*{app:routing_visualization}\quad Task-Dependent Expert Routing}
        \item \hyperref[app:expert_removal]{\ref*{app:expert_removal}\quad Expert Removal}
    \end{itemize}
\end{itemize}

\bigskip

\section{Implementation Details}
\label{app:implementation}

\subsection{Training Data}
\label{app:training_data}

Our training corpus comprises 100 tasks with 1,000 samples per task. Each sample contains an initial state, a task instruction, and a successful reference trajectory. These 100,000 reference samples support latent reasoner training in Stage 1 and video generator adaptation in Stage 2. For localized contrastive-state learning, we additionally generate one negative candidate per sample using VBVR-Wan2.2~\citep{wang2026vbvr} under the same initial state and instruction, yielding 100,000 positive--negative trajectory pairs. Since generated candidates may also satisfy the task, our localization mechanism identifies informative differences within these pairs for supervision. Of the 100 training tasks, 50 are included in the ID evaluation using held-out instances. The 50 OOD evaluation tasks are absent from the training corpus.

For expert pretraining, we first define the reasoning skill associated with each of the seven experts and manually select seven representative anchor tasks for that skill from the same corpus. The task subsets are disjoint across experts, yielding 49 anchor tasks and 49,000 reference samples in total. Each expert is pretrained on its corresponding subset to establish an initial skill specialization. Subsequent router training and joint refinement cover all 100 tasks without additional task-to-expert annotations.

\subsection{Architecture and Training Procedure}
\label{app:training_procedure}

\paragraph{Model initialization.}
We initialize the visual encoder from the V-JEPA 2.1 ViT-L~\citep{murlabadia2026vjepa21} checkpoint \texttt{vjepa2\_1\_vitl\_dist\_vitG\_384.pt}, loading its \texttt{ema\_encoder} weights. The encoder remains frozen throughout both training stages. The state-token pooler and six-layer task-relevant latent predictor are trained for our framework rather than initialized from the pretrained V-JEPA predictor. For Stage 2, we initialize the video generator from VBVR-Wan2.2 and adapt its high-noise branch, while retaining the low-noise branch.

\paragraph{Expert architecture and routing.}
Seven bottleneck-adapter experts are inserted into the final two predictor blocks. Their designated skills are visual discrimination (VIS), numerical reasoning (NUM), object identity and persistence (IDN), spatial reasoning and planning (SPA), physical dynamics (DYN), compositional reasoning (CMP), and abstract rule inference (ABS). Routing uses $\operatorname{Entmax}_{1.5}$~\citep{peters2019sparse} and the objectives in Section~\ref{sec:experts}.

\paragraph{Stage 1: Optimization settings.}
We use a history window of $H=4$ states and represent each state with $M=8$ tokens of dimension $d=256$. The pooler has $N_h=8$ attention heads. Each rollout contains ten states including the initial state, corresponding to a prediction horizon of $T=9$ in Equation~\ref{eq:framework}. Early stopping is disabled. The two-step prediction and VISReg weights in Equation~\ref{eq:prediction_loss} are $\lambda_2=2.0$ and $\lambda_{\mathrm{reg}}=10^{-4}$, respectively. The expert residual coefficient in Equation~\ref{eq:expert_prediction} is fixed to $\alpha=1$; the implementation adds the routed expert output directly to the shared feed-forward output, without learning or scheduling a separate residual scale.

Latent reasoner training consists of base training, expert pretraining, router initialization, and two successive joint refinement phases. All phases use AdamW with $\beta=(0.9,0.95)$, $\epsilon=10^{-8}$, and an effective global batch size of 512. Learning rates follow a linear warmup followed by cosine decay. The V-JEPA encoder remains frozen in every phase.

\paragraph{LCL hyperparameters.}
For localized contrastive-state learning in Section~\ref{sec:counterfactual}, we use the quantile level $q=0.75$ in Equation~\ref{eq:localized_discrepancy} and state-weight thresholds $\tau_{\min}=0.1$ and $\tau_{\mathrm{full}}=0.2$. The numerical stability term for discrepancy normalization in Equation~\ref{eq:token_localization} is $\epsilon=10^{-6}$. The margin and loss weight in Equation~\ref{eq:counterfactual_loss} are $\mu=0.05$ and $\lambda_{\mathrm{LCL}}=0.05$, respectively.

\paragraph{Base latent reasoner training.}
We train the predictor with the one- and two-step objectives, VISReg regularization~\citep{wu2026visreg}, and localized contrastive-state learning (LCL) defined in Section~\ref{sec:counterfactual}. We first train the shared latent reasoner for 20 epochs (3,720 optimizer updates). The learning rates are $10^{-5}$ for the predictor and other trainable modules, $3\times10^{-5}$ for the state-token pooler, and $3\times10^{-6}$ for the text-conditioning projection. We use weight decay of $10^{-3}$ and 200 warmup updates.

\paragraph{Anchor-task expert pretraining.}
Starting from the trained shared reasoner, we specialize the seven experts on their respective anchor-task subsets. For each subset, only the corresponding expert adapters are activated and optimized; the shared predictor, pooler, and other modules remain frozen. Expert pretraining runs for 20 epochs with a learning rate of $10^{-5}$, weight decay of $10^{-3}$, and 200 warmup updates. This phase establishes skill-specific specializations before learning how to route heterogeneous tasks.

\paragraph{Prediction-driven router initialization.}
We then freeze the expert adapters and shared modules and optimize only the router on the multi-task training split. Routing is learned through the latent prediction objective, without task-to-expert labels or an auxiliary load-balancing loss. We use a learning rate of $10^{-4}$, zero weight decay, and 200 warmup updates. The cosine schedule is configured for 20 epochs; the checkpoint used to initialize joint refinement is selected after four epochs (744 optimizer updates).

\paragraph{Joint refinement.}
We first jointly optimize the router and expert adapters for two epochs (372 optimizer updates), using learning rates of $2\times10^{-5}$ and $10^{-5}$, respectively, while keeping the shared modules frozen.

We subsequently unfreeze the predictor's attention layers, shared feed-forward layers, and state-token pooler for three additional epochs (558 optimizer updates). The learning rates are $10^{-5}$ for the router, $5\times10^{-6}$ for expert adapters, $5\times10^{-7}$ for both attention and shared feed-forward layers, and $2\times10^{-7}$ for the pooler. Both refinement phases use zero weight decay and 25 warmup updates. This progressive refinement first coordinates expert selection and specialization, then adapts the shared latent representation to support them.

\paragraph{Stage 2: Video generator adaptation.}
We freeze the trained latent reasoner and optimize the video generator using the flow-matching objective~\citep{lipman2023flowmatching} on reference videos, conditioned on the initial state, instruction, and predicted latent trajectory. Only the LoRA parameters~\citep{hu2022lora} and trajectory-conditioning modules in the adapted Wan2.2 generator branch~\citep{wan2025wan22} are optimized. The trajectory-conditioning interface (TCI) comprises the semantic memory adapter and rollout modulation adapter described in Section~\ref{sec:overview}. The former supplies rollout features through gated cross-attention; the latter produces offsets for the DiT modulation parameters.

We use AdamW with $\beta=(0.9,0.999)$, $\epsilon=10^{-8}$, and weight decay of 0.01. The learning rates are $10^{-5}$ for LoRA and $10^{-4}$ for the trajectory-conditioning modules, with 50 warmup updates followed by constant learning rates. Training runs for 2,000 optimizer updates with an effective global batch size of 64. We use rank-32 LoRA on the attention projections and feed-forward linear layers, and clip the gradient norm to 1.0.

\subsection{Eight-GPU Reproduction Configuration}
\label{app:configuration}

For reproduction on eight H100 GPUs, the Stage 1 effective batch size of 512 can be obtained using a per-GPU batch size of 32 with two gradient accumulation steps for base training, or a per-GPU batch size of eight with eight accumulation steps for expert pretraining. Router initialization and both joint refinement phases use a per-GPU batch size of one with 64 accumulation steps. Stage 2 uses a per-GPU batch size of one with eight accumulation steps, yielding an effective batch size of 64.


\subsection{Parameter Overhead}
\label{app:parameter_overhead}

\methodname{} is initialized from VBVR-Wan2.2 and inherits its 306.708M LoRA parameters across the two generator branches (153.354M per branch). The underlying Wan2.2 DiT contains 28.578B parameters, with approximately 14.289B per branch. Our latent predictor contains 38.093M parameters, including the SR-MoE experts and router. The trajectory-conditioning interface adds 17.181M parameters: 11.807M for the semantic memory adapter and 5.374M for the rollout modulation adapter. Relative to VBVR-Wan2.2, the predictor and conditioning modules therefore add 55.274M parameters, equivalent to approximately 0.19\% of the underlying 28.578B-parameter DiT. Relative to the original Wan2.2 DiT, the inherited LoRA parameters and these new modules together account for 361.982M additional parameters, approximately 1.27\% of the DiT parameter count.

As shown in Table~\ref{tab:main_results}, \methodname{} improves the overall score from 50.3\% to 56.0\% over VBVR-Wan2.2 with only 55.274M additional predictor and conditioning parameters beyond that baseline; the inherited LoRA parameters are already present in VBVR-Wan2.2. This small overhead supports a design that emphasizes informative contrastive supervision through LCL and adaptive skill-expert composition through SR-MoE, rather than substantial model expansion.

\section{Additional Ablation Studies}
\label{app:ablations}
\subsection{Training Strategies}
\label{app:training_strategies}

\begin{table}[h]
    \centering
    \small
    \setlength{\tabcolsep}{7pt}
    \renewcommand{\arraystretch}{1.05}

    \caption{
        \textbf{Ablation of the training strategy.}
We compare tuning the trajectory-conditioning interface (TCI),
LoRA parameters, or both, with VBVR-Wan2.2 as the baseline.
ID and OOD task-success scores are reported on VBVR-Pro-Bench
as percentages; higher is better.
    }
    \label{tab:training_ablation}

    \begin{tabular}{lcccc}
        \toprule
        \multirow{2}{*}{\textbf{Training strategy}}
        & \multicolumn{2}{c}{\textbf{Trainable modules}}
        & \multicolumn{2}{c}{\textbf{Task success (\%)}} \\
        \cmidrule(lr){2-3}
        \cmidrule(lr){4-5}

        & \textbf{TCI}
        & \textbf{LoRA}
        & \textbf{ID} $\uparrow$
        & \textbf{OOD} $\uparrow$ \\
        \midrule

        VBVR-Wan2.2
        & \xmarkg
        & \xmarkg
        & 55.4
        & 45.1 \\

        \addlinespace[2pt]

        LoRA-only tuning
        & \xmarkg
        & \cmark
        & 57.4
        & 45.5 \\

        TCI-only tuning
        & \cmark
        & \xmarkg
        & 63.1
        & 43.7 \\

        Joint tuning
        & \cmark
        & \cmark
        & \textbf{65.9}
        & \textbf{46.1} \\

        \bottomrule
    \end{tabular}
\end{table}

Table~\ref{tab:training_ablation} compares LoRA-only, TCI-only, and joint tuning. Joint tuning achieves 65.9\%/46.1\%, exceeding TCI-only tuning by 2.8 percentage points on ID tasks and 2.4 points on OOD tasks. It also exceeds LoRA-only tuning by 8.5 and 0.6 points, respectively. Training the conditioning interface alone is therefore sufficient to improve ID performance in this comparison, whereas the strongest scores on both splits are obtained when the interface and renderer LoRA parameters are adapted together.


\subsection{Expert Routing Strategies}
\label{app:routing}

\begin{figure}[htbp]
    \centering
    \includegraphics[width=\textwidth]{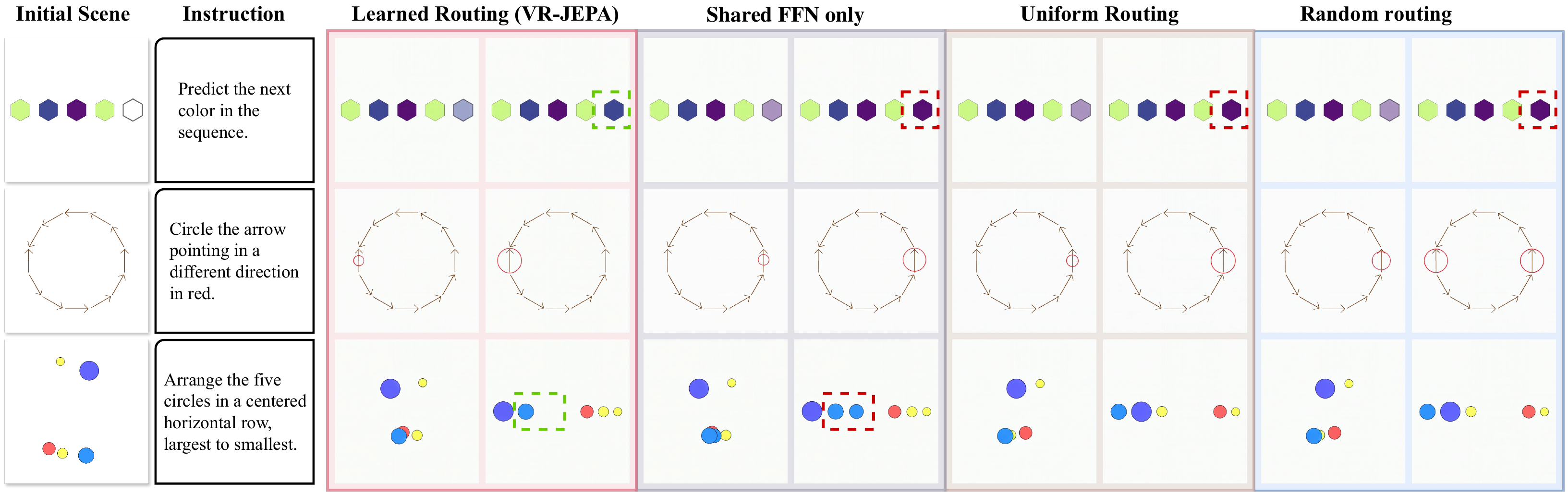}
    \caption{\textbf{Qualitative comparison of SR-MoE routing strategies.} We compare video reasoning results with learned routing, shared FFN only, uniform routing, and random routing. The shared-FFN-only variant disables expert branches at inference. Disrupting learned expert coordination degrades reasoning performance in the illustrated examples. Green and red dashed boxes highlight correct and incorrect results, respectively.}
    \label{fig:sr_moe_ablation}
\end{figure}

\begin{table}[h]
    \centering
    \small
    \setlength{\tabcolsep}{10pt}
    \renewcommand{\arraystretch}{1.08}

    \caption{
        \textbf{Ablation of the SR-MoE routing strategy.}
        We compare learned routing with uniform routing, random routing,
and a shared-FFN-only variant to assess the contribution of
adaptive expert composition. Shared FFN only disables the expert branches of the trained full model at inference, retaining the shared FFNs.
ID and OOD task-success scores are reported on VBVR-Pro-Bench
as percentages; higher is better.
    }
    \label{tab:moe_routing}

    \begin{tabular}{lcc}
        \toprule
        \multirow{2}{*}{\textbf{Routing strategy}}
        & \multicolumn{2}{c}{\textbf{Task success(\%)}} \\
        \cmidrule(lr){2-3}

        & \textbf{ID} $\uparrow$
        & \textbf{OOD} $\uparrow$ \\
        \midrule

        Shared FFN only
        & 64.2
        & 44.7 \\

        \midrule

        Uniform routing
        & 63.9
        & 44.7 \\

        Random routing
        & 64.8
        & 44.8 \\

        Learned routing
        & \textbf{65.9}
        & \textbf{46.1} \\

        \bottomrule
    \end{tabular}
\end{table}
Table~\ref{tab:moe_routing} compares learned routing with uniform and random expert combinations, together with a shared-FFN-only reference. The shared-FFN-only variant disables the expert branches of the trained full model at inference; the + LCL variant in Table~\ref{tab:component_ablation} is trained without SR-MoE. Learned routing achieves the highest ID and OOD scores of 65.9\% and 46.1\%, exceeding uniform routing by 2.0 and 1.4 percentage points and random routing by 1.1 and 1.3 points, respectively. Uniform routing does not improve upon the shared-FFN-only reference, while random routing yields only small increases. These results suggest that the benefit of the expert set depends on how its outputs are combined and support learning the routing weights from the prediction context.
Figure~\ref{fig:sr_moe_ablation} provides qualitative comparisons of the same expert routing strategies.


\subsection{Predicted trajectory}
\label{app:rollout_interventions}
\begin{table}[h]
    \centering
    \small
    \setlength{\tabcolsep}{5pt}
    \renewcommand{\arraystretch}{1.08}

    \caption{
        \textbf{Ablation of latent trajectory guidance.}
        We test whether task success depends on the sample-specific
        temporal content of the rollout, rather than merely the presence
        of an additional conditioning signal.
    }
    \label{tab:rollout_intervention}

    \begin{tabularx}{\linewidth}{
        @{}
        l
        >{\raggedright\arraybackslash}X
        cc
        @{}
    }
        \toprule
        \multirow{2}{*}{\textbf{Rollout condition}}
        & \multirow{2}{*}{\textbf{Intervention}}
        & \multicolumn{2}{c}{\textbf{Task success (\%)}} \\
        \cmidrule(lr){3-4}

        & & \textbf{ID} $\uparrow$
          & \textbf{OOD} $\uparrow$ \\
        \midrule

        Predicted rollout
        & Use the model-predicted trajectory
        & \textbf{65.9}
        & \textbf{46.1} \\

        \midrule

        No conditioning
        & Remove all trajectory conditioning
        & 51.5
        & 44.9 \\

        Repeated initial state
        & Repeat the initial state across all rollout steps
        & 62.5
        & 43.3 \\

        Cross-sample rollout
        & Use the predicted trajectory from another sample
        & 58.5
        & 45.9 \\

        \bottomrule
    \end{tabularx}
\end{table}

Table~\ref{tab:rollout_intervention} examines whether the generator uses the content of the predicted rollout by changing its trajectory input while keeping the observation, instruction, and sampling seed fixed. Removing trajectory conditioning reduces ID performance from 65.9\% to 51.5\%, a drop of 14.4 percentage points. Replacing the predicted rollout with one from another sample yields 58.5\%, a drop of 7.4 points despite retaining the conditioning pathway. \textbf{This comparison indicates that the ID benefit depends in part on the correspondence between the rollout and the current sample, beyond the mere presence of an additional conditioning signal.} The effect is smaller on OOD tasks: removing conditioning lowers the score from 46.1\% to 44.9\%, whereas cross-sample substitution yields 45.9\%, close to the original score. Thus, sample-matched rollout content contributes more clearly to ID performance, consistent with the smaller OOD gains in the main evaluation.







\clearpage
\section{Further Visualization}
\label{app:qualitative}

\subsection{Comparisons with Open-source Models}
\label{app:qualitative_open_source}
\begin{figure}[H]
    \centering
    \includegraphics[width=\linewidth,height=0.76\textheight,keepaspectratio]{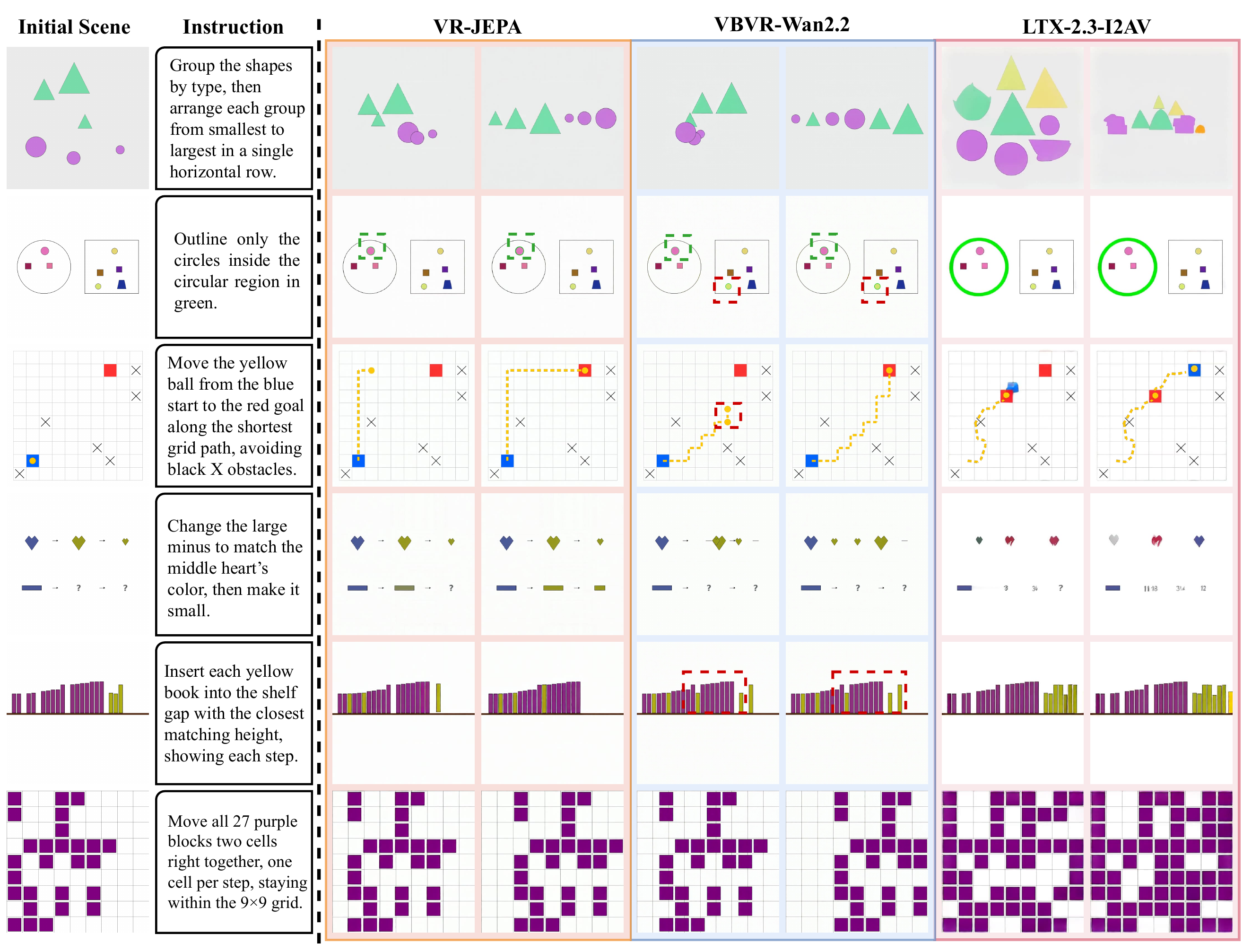}
    \caption{\textbf{Additional qualitative comparisons with open-source models on VBVR-Pro-Bench.} Given the initial state and instruction, we compare the video reasoning results of \methodname{} with VBVR-Wan2.2 and LTX-2.3-I2AV. Green and red dashed boxes highlight correct results and reasoning errors, respectively.}
    \label{fig:qualitative_open_source}
\end{figure}

Figure~\ref{fig:qualitative_open_source} compares \methodname{} with VBVR-Wan2.2 and LTX-2.3-I2AV on additional reasoning tasks. In the shape-sorting example, \methodname{} groups objects by type and orders them by size, while VBVR-Wan2.2 interleaves the groups and LTX-2.3-I2AV changes object attributes. In the selective-outlining example, \methodname{} marks the target circle, whereas the baselines additionally select an outside circle or outline the enclosing region, illustrating errors in applying spatial selection constraints.

\clearpage
\subsection{Comparisons with Proprietary Video Models}
\label{app:qualitative_closed_source}
\begin{figure}[H]
    \centering
    \includegraphics[width=\linewidth,height=0.76\textheight,keepaspectratio]{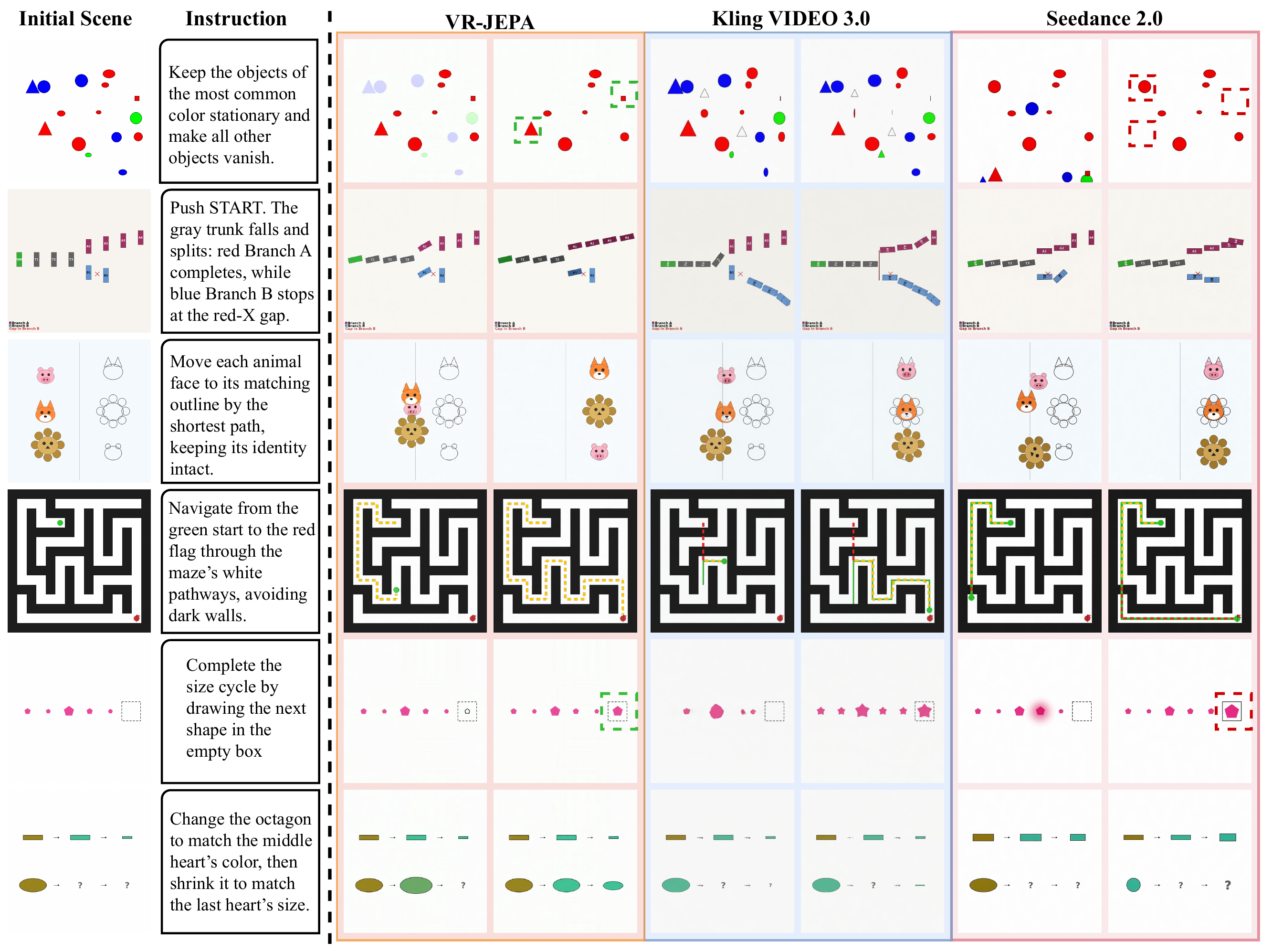}
    \caption{\textbf{Additional qualitative comparisons with proprietary video models on VBVR-Pro-Bench.} Given the initial state and instruction, we compare the video reasoning results of \methodname{} with Kling VIDEO 3.0 and Seedance 2.0. Green and red dashed boxes highlight correct results and reasoning errors, respectively.}
    \label{fig:qualitative_closed_source}
\end{figure}

Figure~\ref{fig:qualitative_closed_source} compares \methodname{} with Kling VIDEO 3.0 and Seedance 2.0. In the domino example, \methodname{} completes the red branch while stopping the blue branch at the marked gap, whereas both baselines propagate beyond it. In the animal-matching example, \methodname{} places each face in its corresponding outline, while both baselines mismatch the identities and destinations. These examples illustrate better adherence to interaction rules and object correspondence.

\clearpage
\subsection{Visual Representation Space Ablation}
\label{app:representation_ablation}
\begin{figure}[H]
    \centering
    \includegraphics[width=\linewidth]{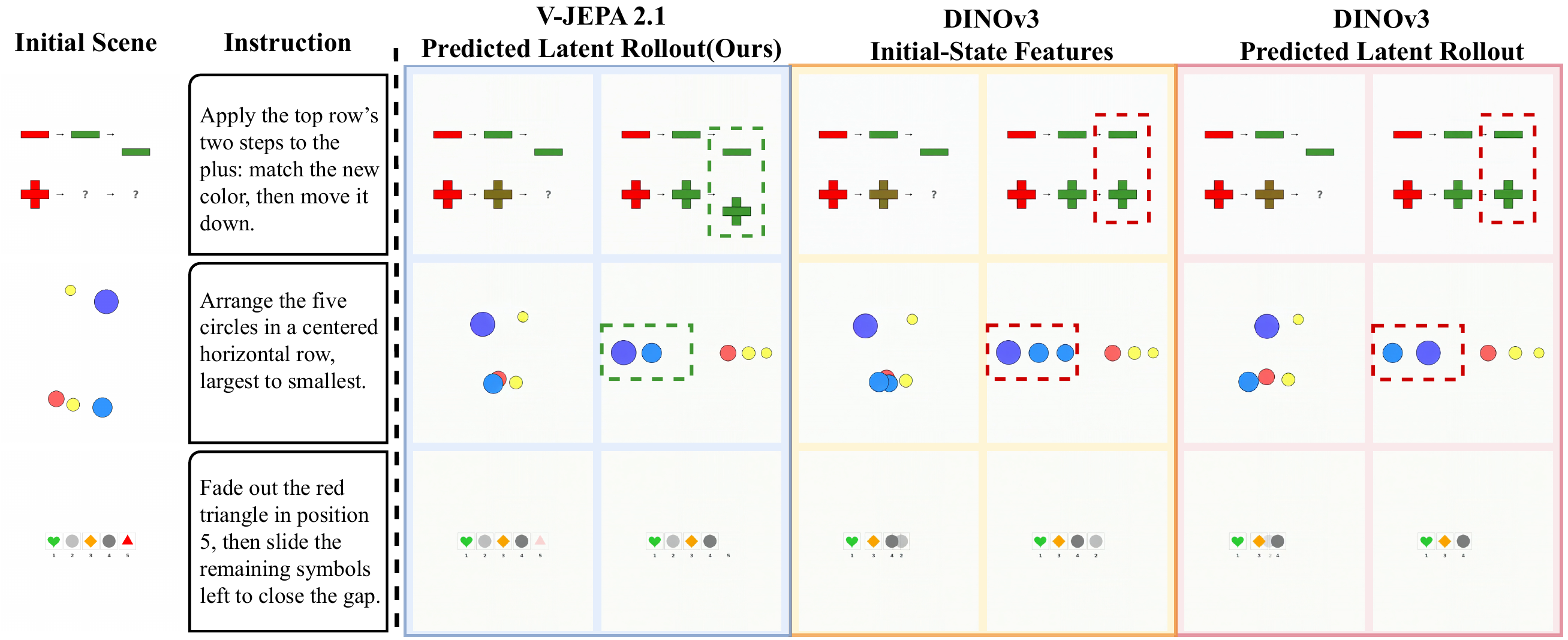}
    \caption{\textbf{Qualitative comparison of visual representations on VBVR-Pro-Bench.} Given the initial state and instruction, we compare video reasoning results guided by, from left to right: (1) predicted latent rollouts in V-JEPA 2.1 space (\methodname{}), (2) initial-state DINOv3 features, and (3) predicted latent rollouts in DINOv3 space. Green and red dashed boxes highlight correct results and reasoning errors, respectively.}
    \label{fig:representation_ablation}
\end{figure}

Figure~\ref{fig:representation_ablation} complements Table~\ref{tab:representation_ablation} by comparing video reasoning results under the three visual representation configurations.

\clearpage
\subsection{Comparison of latent trajectory guidance}
\label{app:rollout_intervention}
\begin{figure}[t]
    \centering
    \includegraphics[width=\textwidth]{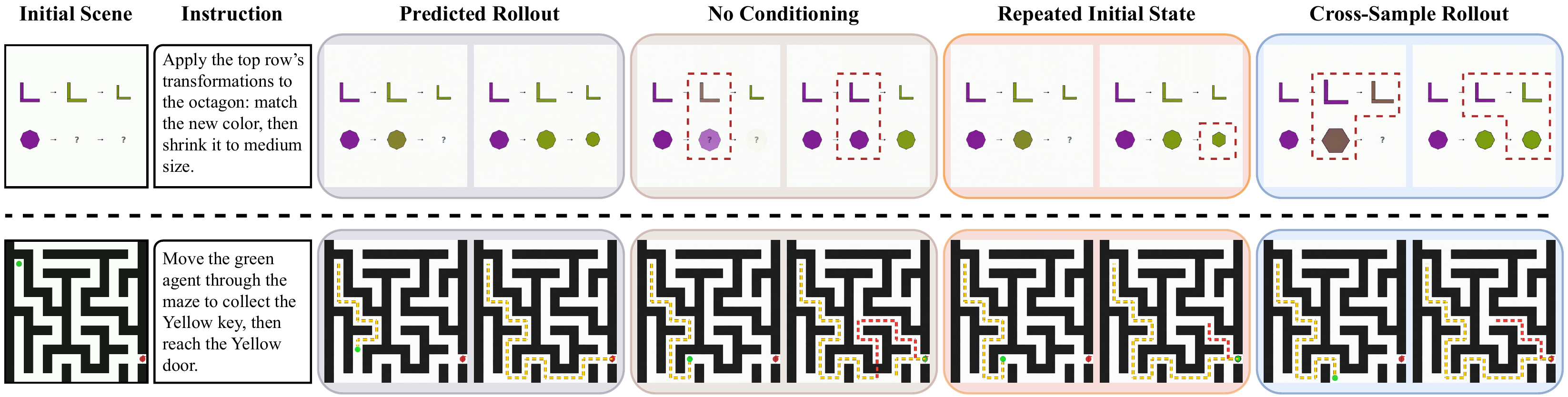}
    \caption{\textbf{Qualitative ablation of latent trajectory guidance.} We compare video reasoning results using the predicted rollout with three interventions: removing trajectory conditioning, repeating the initial state across rollout steps, and substituting a rollout from another sample. Red dashed boxes highlight reasoning errors.}
    \label{fig:rollout_intervention}
\end{figure}

Figure~\ref{fig:rollout_intervention} compares generations conditioned on the predicted rollout with three interventions: removing trajectory conditioning, repeating the initial state, and substituting another sample's rollout. In the upper example, predicted-rollout conditioning more closely follows the reference sequence of shape and attribute changes, whereas the interventions introduce mismatches in intermediate or final states. In the maze example, the predicted-rollout result more closely follows the reference path, while the intervened results deviate at the locations marked in red. Removing conditioning tests the presence of trajectory guidance; repeating the initial state tests whether its temporal evolution matters; and substituting another sample's rollout tests the importance of instance-specific content. These examples illustrate the different failure modes, while the quantitative intervention results are needed to assess how consistently they occur.

\clearpage
\subsection{Task-Dependent Expert Routing}
\label{app:routing_visualization}

\begin{figure}[H]
    \centering
    \includegraphics[width=\textwidth]{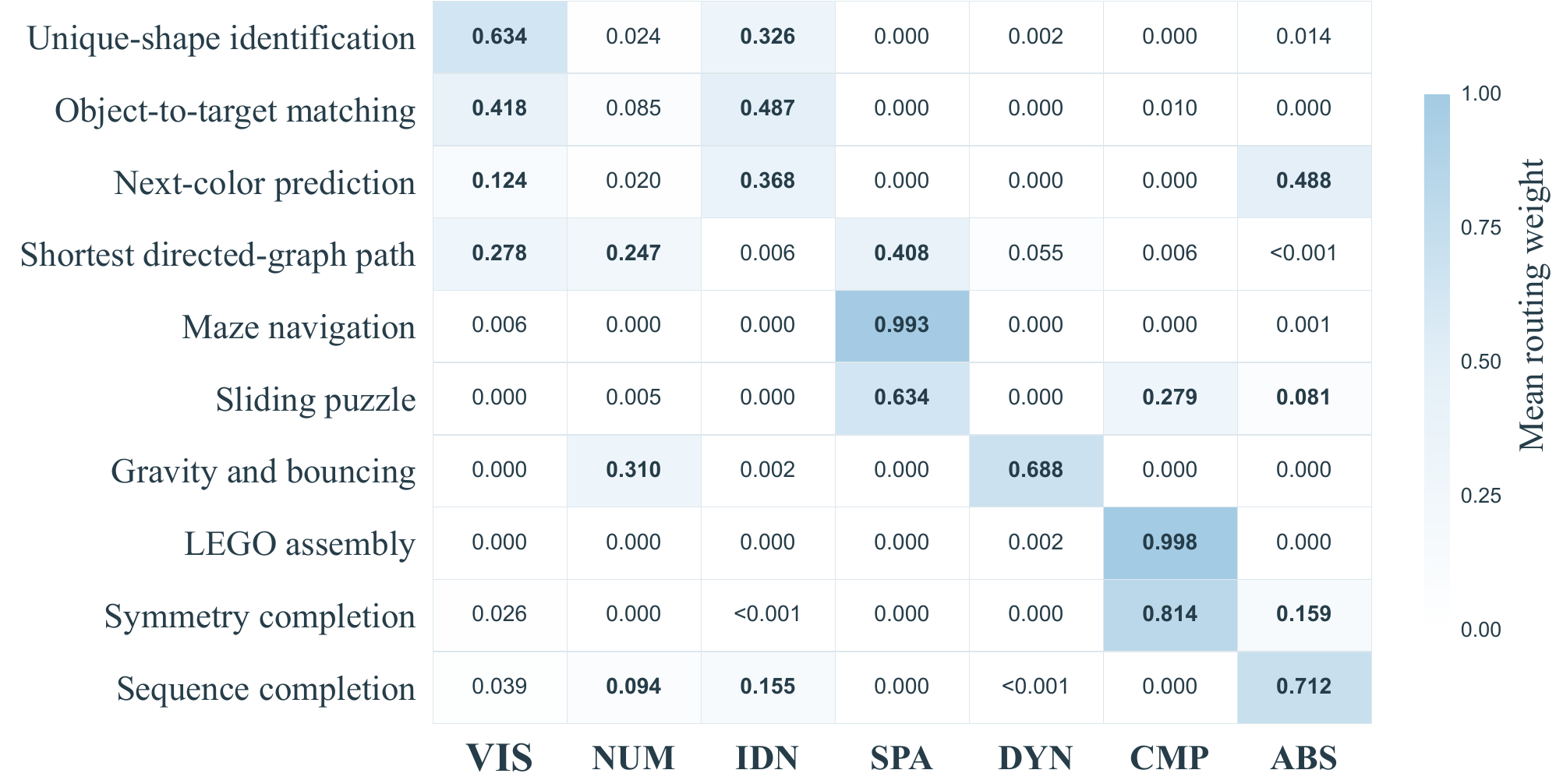}
    \caption{\textbf{Task-dependent expert routing.} Mean routing weights for ten tasks, averaged across the two SR-MoE predictor layers (layer4 and layer5). Rows denote tasks and columns denote the seven skill-specific experts; darker cells indicate larger weights. The distributions show both concentrated expert selection and allocations spread across multiple experts.}
    \label{fig:routing_heatmap}
\end{figure}

Figure~\ref{fig:routing_heatmap} visualizes the mean routing weights for ten tasks, averaged across layer4 and layer5, the two SR-MoE blocks of the latent predictor. The task-dependent distributions are broadly consistent with the experts' designated skills. Maze navigation and LEGO assembly assign nearly all routing weight to SPA (0.993) and CMP (0.998), respectively, while gravity and bouncing favors DYN (0.688). Unique-shape identification assigns its largest weight to VIS, object-to-target matching to IDN, and next-color prediction and sequence completion to ABS.

Other tasks distribute substantial average weight across several experts. Sliding puzzle allocates weight primarily to SPA (0.634) and CMP (0.279), while shortest directed-graph path draws on SPA (0.408), VIS (0.278), and NUM (0.247). These patterns are consistent with task-dependent expert composition rather than a fixed assignment of every task to a single expert. The averages summarize expert utilization rather than establish each expert's causal contribution; the removal analysis in Section~\ref{app:expert_removal} provides complementary evidence by examining the effects of suppressing a selected expert.

\clearpage
\subsection{Skill-specific expert removal}
\label{app:expert_removal}
\label{sec:expert_removal_visualization}

\begin{figure}[H]
    \centering
    \includegraphics[width=\textwidth,height=0.72\textheight,keepaspectratio]{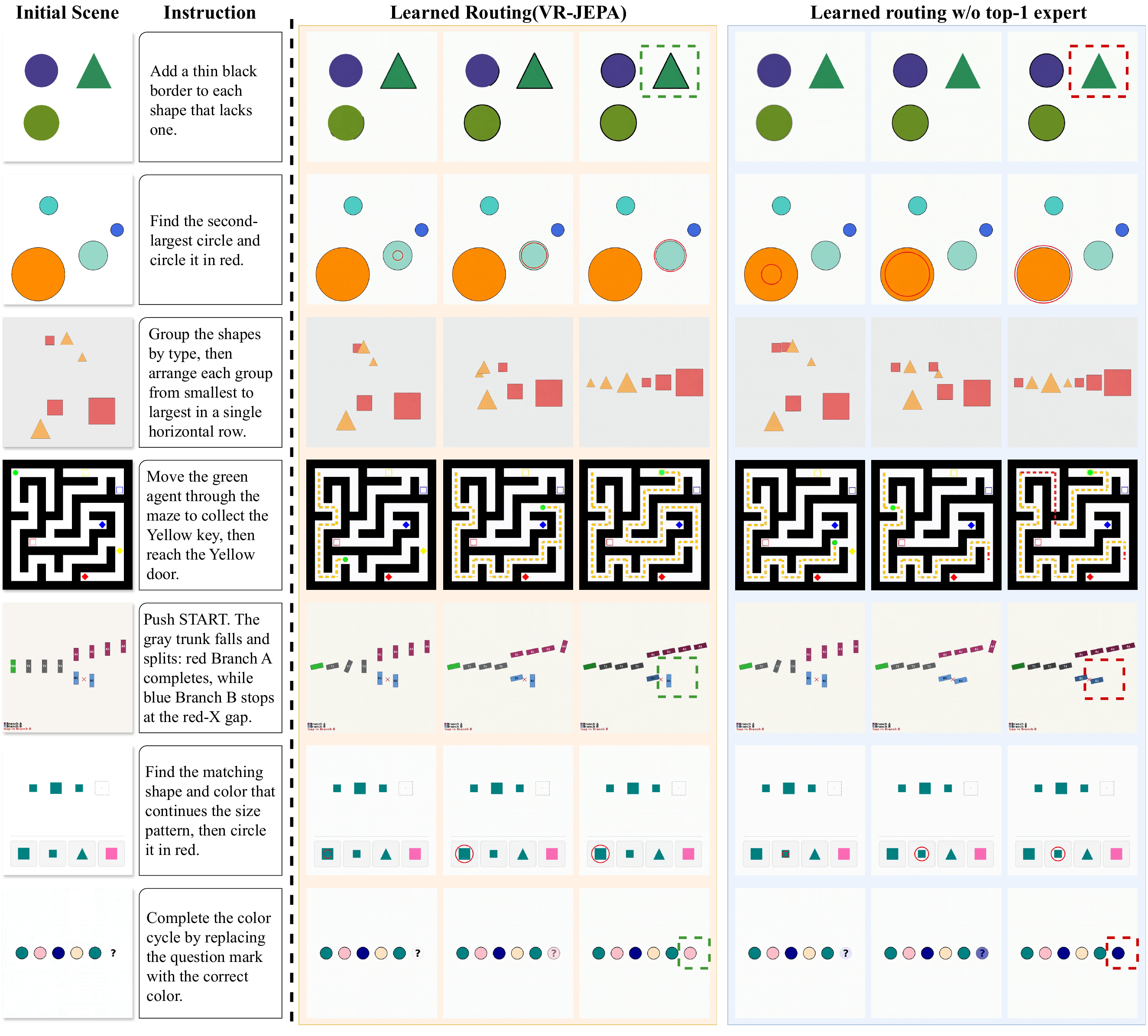}
    \caption{\textbf{Probing expert specialization through skill removal.} We remove the highest-weight expert for each of seven examples and examine whether the resulting errors reflect the loss of its associated reasoning skill. From top to bottom, the removed experts correspond to visual discrimination, numerical reasoning, object identity and persistence, spatial reasoning and planning, physical dynamics, compositional reasoning, and abstract rule inference. Green and red dashed boxes highlight correct results and reasoning errors, respectively.}
    \label{fig:drop_top1_expert}
\end{figure}

Figure~\ref{fig:drop_top1_expert} compares learned routing with removal of the highest-weight expert in seven selected examples, ordered by the designated skills VIS, NUM, IDN, SPA, DYN, CMP, and ABS. The interventions produce several distinct errors. Border completion leaves a shape unoutlined; ordinal selection marks the largest circle instead of the second-largest; and shape sorting duplicates a small red square. The maze output includes an invalid path, and the domino output propagates beyond the marked gap. The final two examples select the wrong size in a shape-color sequence and the wrong color in a repeating cycle.

These errors are consistent with the reasoning operations associated with the removed experts and provide qualitative evidence of differentiated contributions. Because the experiment removes the highest-weight expert in selected examples, it does not establish an exclusive mapping between each expert and one skill or separate skill specificity from general sensitivity to removing a strongly weighted expert.

\clearpage

\end{document}